\documentclass[letterpaper, 10pt, journal, twoside]{IEEEtran}
\def\figurespace{\vspace{-3ex}}

\ifCLASSINFOpdf
  \usepackage[pdftex]{graphicx}
\else
\fi
\usepackage{amsmath}
\usepackage{amssymb}

\usepackage[font=small]{subcaption}
\usepackage[font=small,skip=3pt,tableposition=top]{caption}
\usepackage{adjustbox}
\usepackage{booktabs}
\usepackage{xcolor}

\usepackage{url}

\begin{document}
%
\title{\LARGE \bf
Learning Composable Behavior Embeddings for Long-horizon \\ Visual Navigation
}

%
%
%

\author{Xiangyun Meng, Yu Xiang and Dieter Fox
\thanks{Manuscript received Oct 15, 2020; Revised Dec 30, 2020; Accepted Jan 28, 2021.}
\thanks{This paper was recommended for publication by Editor Eric Marchand upon evaluation of the Associate Editor and Reviewers’ comments.}
\thanks{This work was funded in part by ONR grant 63-6094 and by the Honda Curious Minded Sponsored Research Agreement.}
\thanks{Xiangyun Meng and Dieter Fox are with the Paul G. Allen School of Computer Science \& Engineering, University of Washington, Seattle, WA 98195, USA {\tt\small\{xiangyun, fox\}@cs.washington.edu}}
\thanks{Yu Xiang and Dieter Fox are with NVIDIA, Seattle, WA 98105, USA {\tt\small \{ yux, dieterf\}@nvidia.com }}
\thanks{Digital Object Identifier (DOI): see top of this page.}
}

\markboth{IEEE Robotics and Automation Letters. Preprint Version. Accepted Jan, 2021}{Meng \MakeLowercase{\textit{et al.}}: Learning Composable Behavior Embeddings for Long-horizon Visual Navigation}

%


\IEEEpubid{\makebox[\columnwidth]{\copyright{}2021 IEEE} \hspace{\columnsep}\makebox[\columnwidth]{ }}



\maketitle

\begin{abstract}
Learning high-level navigation behaviors has important implications: it enables robots to build compact visual memory for repeating demonstrations and to build sparse topological maps for planning in novel environments. Existing approaches only learn discrete, short-horizon behaviors. These standalone behaviors usually assume a discrete action space with simple robot dynamics, thus they cannot capture the intricacy and complexity of real-world trajectories. To this end, we propose \emph{Composable Behavior Embedding} (CBE), a continuous behavior representation for long-horizon visual navigation. CBE is learned in an end-to-end fashion; it effectively captures path geometry and is robust to unseen obstacles. We show that CBE can be used to performing memory-efficient path following and topological mapping, saving more than an order of magnitude of memory than behavior-less approaches. For more information, visit \color{blue}{\url{ https://homes.cs.washington.edu/~xiangyun/ral21/}}
\end{abstract}


%
\IEEEpeerreviewmaketitle

\section{Introduction}
%
%
%
%
\IEEEPARstart{H}{umans} spend significant amounts of time navigating between familiar places: grabbing a cup of coffee from the kitchen, going to the printer room to collect papers, or retrieving mails from the mailbox. These navigation routines are executed with little conscious effort because they are so repetitive that they have almost become part of our muscle memory. This saves cognitive load, allowing us to concentrate on more important tasks. From a robot learning perspective, enabling a robot to perform such navigation routines robustly with minimal guidance is beneficial, because it saves memory, speeds up computation, and opens up opportunities to build sparse visual memory of environments for efficient planning and control.

Learning high-level behaviors or skills for robots has become an important area of research recently. Most existing works focus on synthetic control tasks or fixed workspace manipulation tasks \cite{shankar2019discovering, noseworthy2020task, lynch2020learning, krishnan2017ddco}, where environments are fully observable and a robot can be position-controlled. This is hardly applicable to egocentric visual navigation, where environments are partially observable, ground truth states are unavailable, and robots may have non-holonomic kinematic constraints. Because of this, most recent works use predefined behaviors \cite{chen2019graphnav, trigonisnapnav, roh2020conditional}, with a few attempts on unsupervised or self-supervised behavior learning \cite{kumar2020learning, swedish2018deep, gopalansimultaneously}. However, these behaviors are usually short-horizon (e.g., ``turn left''), discrete, and cannot follow precise specifications (e.g., distance to go or angle to turn). Due to these limitations, they are not able to encode \emph{complex and long-horizon} behaviors in a general fashion, such as when following an instruction ``go towards northeast by about 5 meters and then turn right to follow the hallway till the end''. This limits their applicability in building sparse visual memory for downstream navigation tasks.

\begin{figure}[t]
    \centering
    \includegraphics[width=0.8\columnwidth]{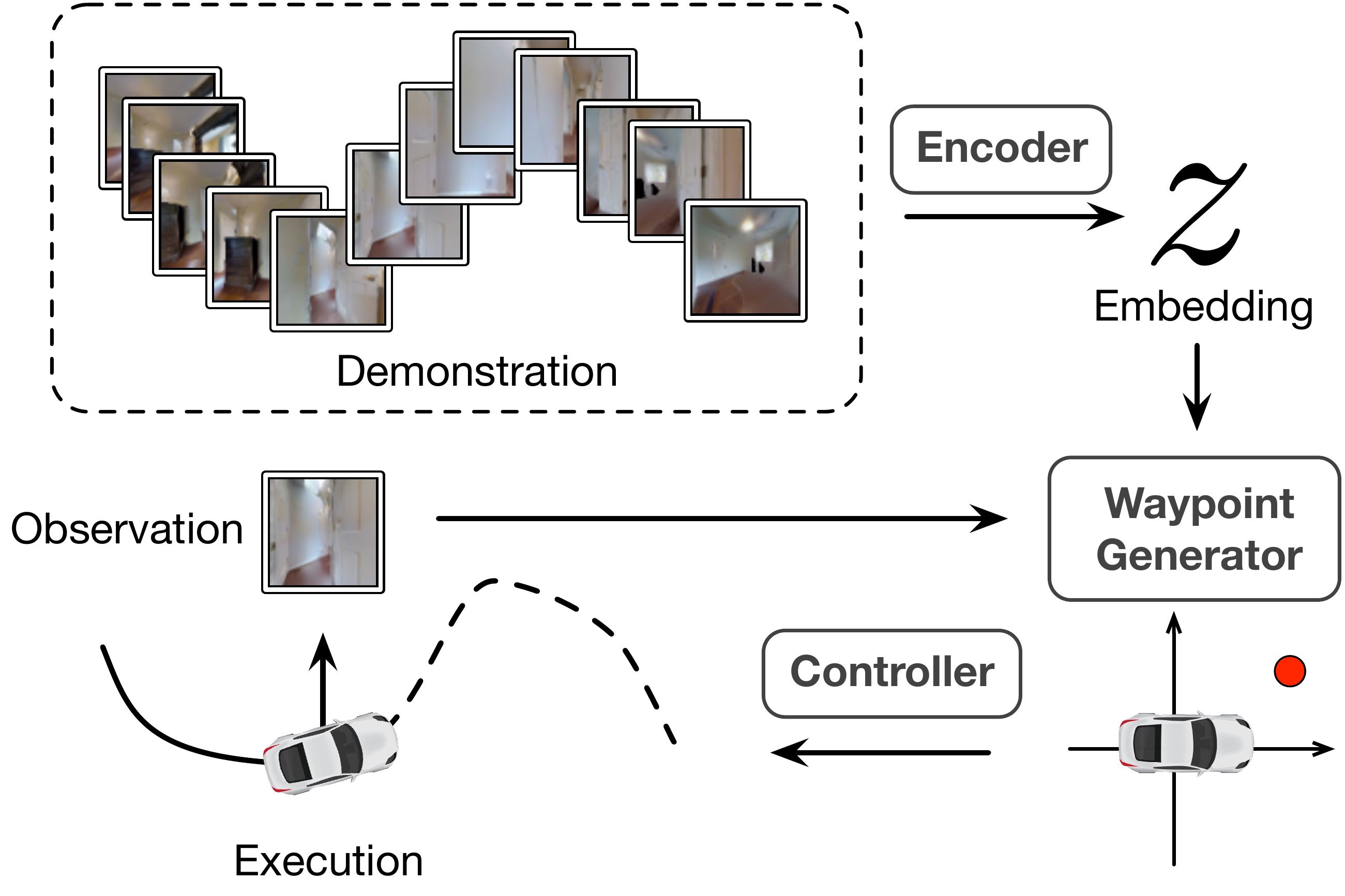}
    \caption{High-level overview of behavior learning and execution. CBE learns to embed image sequences to replicate a demonstrated trajectory using visual, closed-loop waypoint generation and control.}
    \label{fig:overview}
    \figurespace
\end{figure}

To address this problem, we propose \emph{Composable Behavior Embedding}, a robot-agnostic behavior representation for visual navigation (Figure~\ref{fig:overview}). At its core is a behavior encoder that compresses a high-dimensional visual demonstration sequence into a low-dimensional embedding. During execution, a waypoint generator is conditioned on the embedding and current observation to generate local waypoints for a low-level controller to replicate the demonstration. The embeddings are learned in an end-to-end fashion by minimizing the waypoint reconstruction loss. It effectively learns to extract path geometry from demonstrations, making it generalize extremely well to novel environments. 

CBE has two desired properties: i) it is compact. The embedding is only 32-dimensional, allowing a robot to build visual memory an order of magnitude smaller than existing approaches \cite{meng2019scaling, kumar2018visual}, and ii) it is composable. A robot can robustly follow a long path via behavior segmentation, or combine behaviors from multiple demonstrations to perform goal-directed navigation tasks.

SLAM \cite{orbslam, furgale2010visual} can be a strong alternative for building visual memory. CBE has several advantages over SLAM: i) CBE is more than 10\emph{x} efficient at encoding demonstrations than SLAM; ii) CBE works with low-resolution images where SLAM breaks down, allowing it to be deployed on miniature robots without high-quality cameras; iii) CBE has a simpler design with few tuning parameters and is end-to-end trainable. Hence, CBE is a more attractive approach towards building a robust and efficient learning-based visual navigation system.

We show how the embeddings generated by CBE enable a non-holonomic robot to reach goals more than 150 time steps away with no intermediate guidance, even when unseen obstacles are present. We further illustrate how the learned embeddings can be applied to two downstream tasks: one-shot trajectory following and topological mapping. We show that with the learned embeddings we can build visual memory an order of magnitude smaller than existing approaches for these downstream tasks. We conduct detailed quantitative and qualitative analysis to verify our design decisions and how it is compared to a variety of baselines.

\section{Related Work}
\label{sec:related_work}

\textbf{Visual Navigation.} 
Classical navigation systems rely on building a metric map from laser scans or visual images for robust state estimation, planning, and control~\cite{ProbRob05, furgale2010visual}. Recent advances in visual navigation move towards non-metric, learning-based methods, such as short-horizon goal-directed navigation \cite{pathak2018zero}, path following \cite{kumar2018visual, hirose2019deep, trigonisnapnav}, or building a cognitive mapping system for planning \cite{savinov2018semi,chaplot2020neural, meng2019scaling, gupta2017cognitive}. Solving long-horizon navigation tasks requires some form of visual memory \cite{savinov2018semi, meng2019scaling, fang2019scene, kumar2018visual, hirose2019deep}. Due to visual occlusion, dense observations have to be stored, making it difficult to scale to large environments. Our main contribution is to learn a compact embedding so that a robot only stores a sparse set of visual features. These embeddings serve as sparse visual memory for diverse downstream tasks, such as path following and topological mapping.

\textbf{Learning from Demonstrations.} Perhaps the most direct approach to learn from demonstrations is imitation learning \cite{longrange_neural_nav,codevilla2018end}. Imitation learning learns fixed policies that are hard to generalize to novel tasks. Recent works learn latent distributions to encode a diverse skill set \cite{mandlekar2020iris, mandlekar2020learning, shankar2019discovering, lynch2020learning}. These works focus on manipulation tasks in a fully observable workspace, and hence they cannot generalize to novel, partially observable environments as in indoor navigation. Contrary to existing works that hardcode environments into the skills, we learn a shared behavior encoder, allowing a robot to adapt to new environments quickly.

\textbf{Unsupervised Skill Learning.} Learning high-level skills helps to solve long-horizon tasks more effectively. However, most works on skill learning assume fully observable state spaces in known environments \cite{co2018self, sharma2019dynamics, kipf2019compile, krishnan2017ddco, lioutikov2017learning}. This is not applicable in egocentric visual navigation, where environments are partially observable and no ground truth robot state is available. 
So far only discrete, short-horizon navigation skills can be learned \cite{chaplot2019learning, swedish2018deep}, and these skills have only been used for exploration and point-goal navigation tasks. To the best of our knowledge, we are the first to show that diverse and long-horizon navigation skills can be effectively learned from visual data. Moreover, we show that these skills can serve as building blocks for constructing a sparse persistent spatial memory for navigating in novel environments.

\textbf{Sequence-to-Sequence Models.} Our method is inspired by seq2seq models, which have been widely used in language processing \cite{sutskever2014sequence, Anderson_2018_CVPR} and trajectory prediction \cite{lee2017desire, alahi2016social}. An important distinction is that we save the latent states as part of the visual memory. Moreover, our decoding process generates controls that are conditioned on the current rollout, which is essential for correcting drift and avoiding obstacles.

\section{Composable Behavior Embedding (CBE)}
\label{sec:methods}
\subsection{Overview}

We consider a goal-directed navigation task where a robot needs to navigate from its current location $s$ to a goal $g$. We assume that a demonstration containing a sequence of RGB observations $o_1, o_2, ..., o_T$ connecting $s$ and $g$ is given to the robot. Since the trajectory can be long and complex, intermediate information needs to be memorized to help the robot follow the demonstration \cite{kumar2018visual, hirose2019deep, meng2019scaling}.  

Similar to~\cite{meng2019neural}, our navigation system guides the robot by generating \emph{relative waypoints} that are used by a low-level controller to compute motor commands (e.g., velocity and steering angle).  To follow a demonstrated trajectory, the robot could use visual control to match its observations against the sequence of demonstrated observations as in~\cite{hirose2019deep, meng2019scaling, kumar2018visual}.  However, such an approach is highly memory inefficient, since it requires rather densely stored images.  To overcome this problem, CBE encodes the sequence of visual observations $o_1, o_2, ..., o_T$ into a low-dimensional behavior embedding $z_D$ (Figure~\ref{fig:overview}). During execution, at each time step, a waypoint generator uses $z_D$ and the current observation to produce a waypoint for the low-level controller. Both state and action space are continuous, and the system operates in a closed-loop fashion to correct any drift due to noise and non-holonomic kinematic constraints. Different demonstrations and their executions can be of different lengths.  

For very long and complex trajectories, a single $z$ is insufficient due to compounding errors. To address this problem, we segment long trajectories into sequences of embeddings, interleaved by visual attractors for state calibration. The segmented behaviors are used for solving downstream navigation tasks, which we detail in Sec~\ref{sec:result}.

\begin{figure*}
    \centering
    \includegraphics[width=0.85\textwidth]{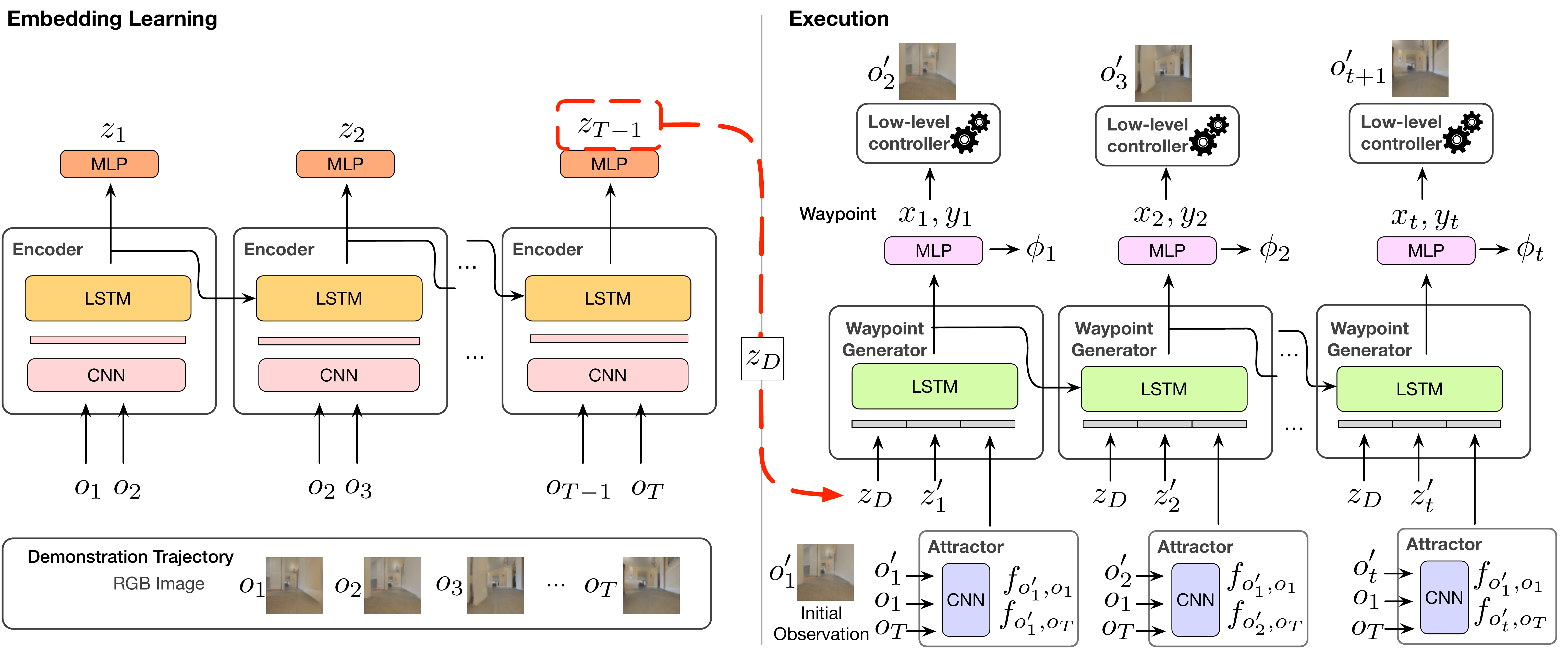}
    \caption{Overview of CBE. The encoder compresses the image sequence $o_1, ..., o_T$ observed during a demonstration into a low dimensional embedding $z_{D}$. This is done via a recurrent LSTM network that inputs pairs of consecutive images, $(o_t, o_{t+1})$, and generates a sequence of latent embeddings, $z_t$, with the final $z_{T-1}$ providing the overall embedding of the demonstration. In the execution phase, the waypoint generator uses the demonstration embedding, $z_{D}$, and the embedding of the images observed so far, $z'_{t}$, to generate the next waypoint, $(x_t, y_t)$, and a measure of the progress thus far $\phi_t$.  The embeddings of the executed trajectory, $z'_{t}$, are computed using the same network as the demonstration encoding.  An additional ``Attractor'' network processes the current image and the first and last images of the demonstration to provide information that helps with the alignment at the beginning and end of the trajectory.  At each time step $t$, the waypoint is sent to the local controller, which moves the robot and provides the image for the next iteration, $o'_{t+1}$.  This process is repeated until the robot reaches the goal $o_T$, indicated by $\phi=1$.}
    \label{fig:arch}
    \figurespace
\end{figure*}

\subsection{Learning Continuous Navigation Behaviors}
The behavior encoder $\mathbb{B}_\text{enc}$ (left half of Figure~\ref{fig:arch}) maps observation streams $o_1, o_2, o_3, ..., o_T$ into a low-dimensional embedding  $z_D = \mathbb{B}_\text{enc}(o_1, o_2, ..., o_T)$. To do so, each pair of adjacent images is input to a CNN that generates a feature vector fed into an LSTM to compute the embedding for each time step. Since the encoder is recurrent, it outputs a sequence of embeddings, where embedding $z_i$ encodes the observed behavior from $o_1$ to $o_{i+1}$. The complete trajectory is encoded into $z_{T-1}$ (i.e., $z_D$).  

Since encoding and execution are only coupled by the embedding, the whole CBE network can be trained end-to-end. Through end-to-end learning, the encoder learns a common behavior manifold (Sec.\ref{sec:result}). The embedding can be extremely low-dimensional (e.g., 32), which significantly saves memory compared to SLAM \cite{orbslam} or other learning-based approaches \cite{kumar2018visual, meng2019scaling}.

\subsection{Behavior-Conditioned Waypoint Generator}
The waypoint generator (right side of Figure~\ref{fig:arch}) executes a behavior while tracking the robot's progress. The robot starts with its initial observation, $o'_1$, which does not have to exactly match the beginning of the demonstration, $o_1$. At every time step $t$, an LSTM  unit takes as input the embedding $z_D$ of the demonstration and the embedding $z'_t$ of the images observed so far (computed using the same encoder network used for demonstration embeddings), along with features provided by an ``Attractor'' network described below.  Using these, the recurrent unit predicts the next local waypoint, $x_t, y_t$, and the current progress, $\phi_t$. The waypoint is input into the robot's low-level controller to generate motor commands.  The low-level controller can be a simple PID controller, or it may support local obstacle avoidance \cite{meng2019neural}.  The \emph{progress indicator} $\phi_t$ provides the fraction of the demonstration the robot has completed at time $t$. It is used as a condition for behavior switching. After receiving  the next observation, $o'_{t+1}$, new attractor features and embedding $z'_{t+1}$ are computed and input to the next LSTM step. This process is repeated until the robot reaches the goal, indicated by $\phi = 1$. 

\textbf{Attractor Network}. During execution, the robot's initial location and orientation may not exactly match the beginning of a demonstration, requiring the robot to align its initial location sufficiently well to follow the demonstrated trajectory. Similarly, to determine when the robot has reached the goal, solely accumulating motion information from the observed images is not accurate enough.  CBE solves these problems via the \emph{attractor} network, which combines the robot's current observation $o'_t$ with $o_1$ and $o_T$ (i.e, attractors) to provide features that can relate the current observation to the beginning and end of the demonstration. The attractor network is a CNN that generates $f_{o'_1, o_1}$ and $f_{o'_t, o_T}$ which are concatenated with the embedding (see Figure~\ref{fig:arch} and \ref{fig:networks}).

\subsection{Long Range Navigation via Behavior Segmentation}
\label{sec:segmentation}
Since behavior embeddings are learned from egocentric observations, compounding error is inevitable, implying that $z$ may not encode a complex long-horizon behavior precisely. We solve this by segmenting a long trajectory into a sequence of behaviors, each of which is specified by its embedding $z_D$ and initial and final observations, $o_1$ and $o_T$, respectively.  Via the attractor features, $o_1$ provides robustness toward noisy locations when starting a behavior, and $o_T$ helps the behavior reach the goal location accurately enough to transition to the next behavior (related to funnels in LQR-Trees~\cite{Tedrake2010}). 

We find fixed-distance segmentation works well in practice (Sec.~\ref{sec:result}). Given an observation sequence $o_1, o_2, ... o_T$, we segment it into equally spaced segments, subject to the constraint that every segment contains no more than $K$ observations, where $K$ is determined by a validation set. Visual attractors are placed at the segmentation boundaries, and two adjacent segments share attractors.

\textbf{Behavior Switching.} When a robot executes a sequence of behaviors, it needs to know when it can safely switch from the current behavior to the next. It makes the switching decision by checking if the progress indicator of executing the current behavior $\phi_\text{current}$ is close to 1 (set to 0.95 in practice). If the condition holds, the robot resets its internal states and starts executing the next behavior $z_\text{next}$.

\subsection{Composing Behaviors from Multiple Demonstrations}
\label{sec:recomposition}

\begin{figure}
    \centering
    \includegraphics[width=0.6\columnwidth]{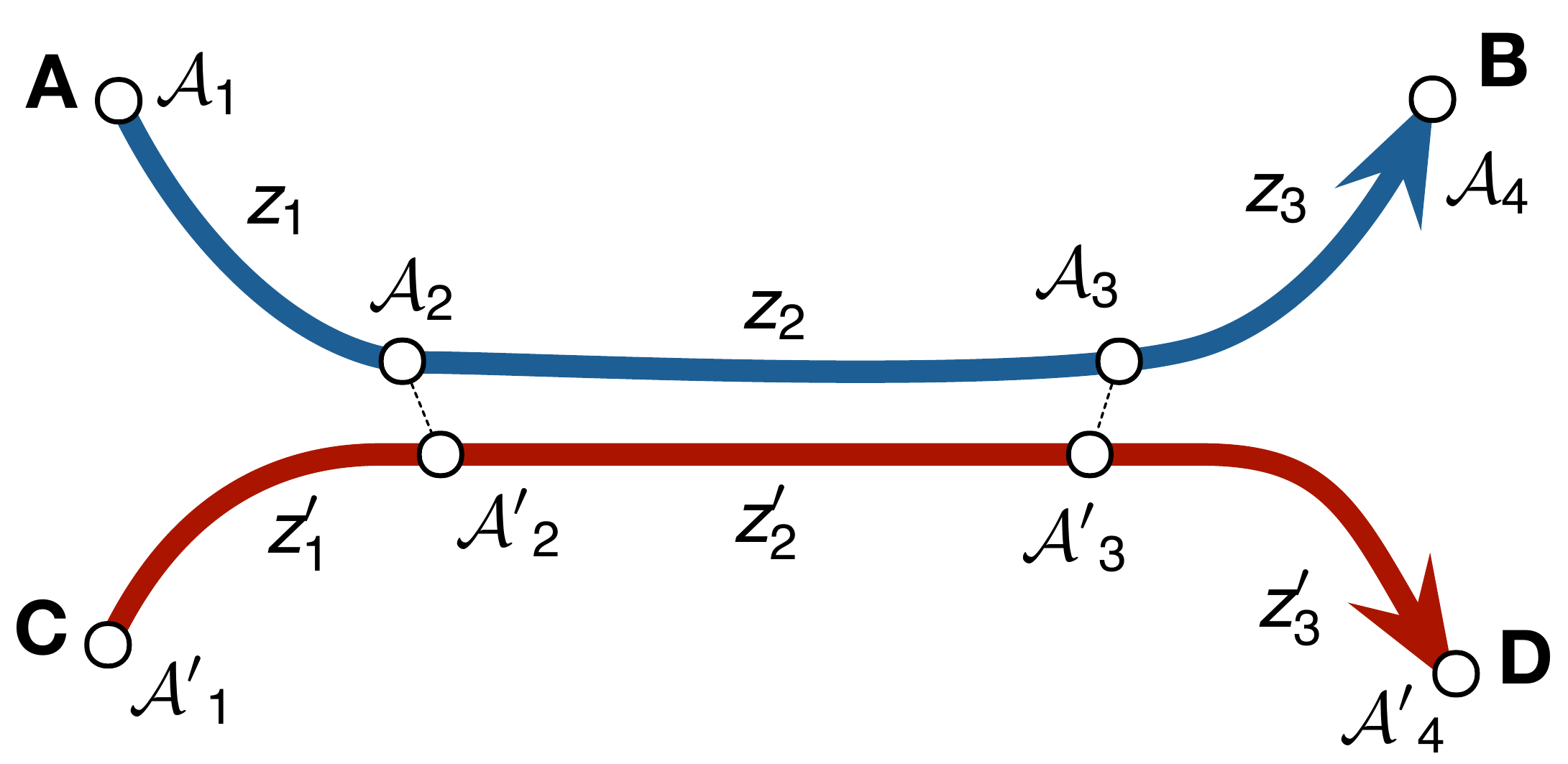}
    \caption{Linking attractors from two demonstrations. $\mathcal{A}_i$, $\mathcal{A}'_j$ are attractors. $z_i$, $z'_j$ are  embeddings between attractors. Dotted lines are connections between attractors that are visually close.}
    \label{fig:recompose_behaviors}
    \figurespace
\end{figure}
 
The segmentation method described in Sec.~\ref{sec:segmentation} can be extended to enable a robot to re-compose behavior segments from multiple demonstrations. In Figure~\ref{fig:recompose_behaviors}, a robot is given demonstrations $A\rightarrow B$ and $C\rightarrow D$. If the attractor  $\mathcal{A}_2$ and $\mathcal{A}'_2$ are close enough, then the robot can execute behaviors $z_1, z'_2, z'_3$ sequentially to go from $A$ to $D$, even though no direct demonstration is available. This also allows us to further compress demonstrations by removing repeated behaviors (e.g., one of $z_2$ and $z'_2$).

\textbf{Learning Choice Points.} Fixed-distance segmentation does not guarantee that visual attractors from different demonstrations are placed at consistent locations, making it difficult to connect demonstrations. To mitigate this, we use a simple algorithm to find spatially consistent attractors. We train a classifier $d_t = \mathbb{C}(o_{t-k+1},...,o_{t})$ that takes the most recent $k$ observations and predicts the next waypoint direction (discretized into 128 bins) using the training dataset. We compute the variance $\sigma_t$ of the directional distribution to measure the uncertainty. Intuitively, $o_t$ with high variance suggests that future trajectories may diverge and thus $o_t$ is usually associated with spatially consistent locations such as intersections and doorways. Given a trajectory $o_1, ..., o_T$, we use $\mathbb{C}$ to compute the directional variances $\sigma_1, ..., \sigma_T$. Then we use a peak finding algorithm to find a set of choice points along a trajectory. While there are more sophisticated methods such as \cite{loquercio2020general} that can potentially find better choice points,  we find this simple approach to be effective (see Sec.~\ref{sec:mapping}).

\section{Implementation Details}
We collected 100k trajectories from 18 large Gibson \cite{xia2020interactive} environments as the training set. The trajectories are generated by a laser-based RMP controller \cite{meng2019neural} driving a non-holonomic car to follow a sequence of local waypoints computed by an A* planner. This controller also serves as the low-level controller for behavior execution. The low-level controller uses laser scans for local obstacle avoidance and in practice it could be replaced with vision-based controllers \cite{hirose2019deep, meng2019neural} at extra computational cost. Simulation runs at 10~Hz. Image resolution is $64\times 64$ with $120^\circ$ field of view. Camera height is set to 1.0~m above the floor. All evaluations are conducted in 5 large unseen Gibson environments. These large scenes are several times the size of an average Gibson scene, hence they are more suitable for evaluating long-horizon navigation performance.

We use a sequence length of 64 with a frame gap uniformly sampled between 0 to 2. Hence the average trajectory length is 128 time steps.  We use the local waypoints in the same training set as supervision, and adopt DAgger \cite{ross2011reduction} for data augmentation. In the DAgger phase, we jitter the robot's initial pose to simulate imperfect alignment and collect rollout trajectories generated by the current model. We then compute the correct waypoints and progress to train the next model. The correct local waypoints are computed by transforming (i.e., rotating and translating) the global ground truth waypoint associated with the closest trajectory sample to the robot. To compute the correct progress, we define the completed path as $o_1, ..., o_k$ where $o_k$ is the closest observation to $o'_t$ (in Euclidean distance). Hence $\phi_t$ is the fraction of completed path length to the total path length. By jittering the robot's pose in the DAgger phase, the robot learns a closed-loop policy that is robust to drift. This also enables the robot to robustly switch to the next behavior segment albeit the initial misalignment and errors in progress estimation by relating current observation to the attractors.

\begin{figure}
    \centering
    \includegraphics[width=0.85\columnwidth]{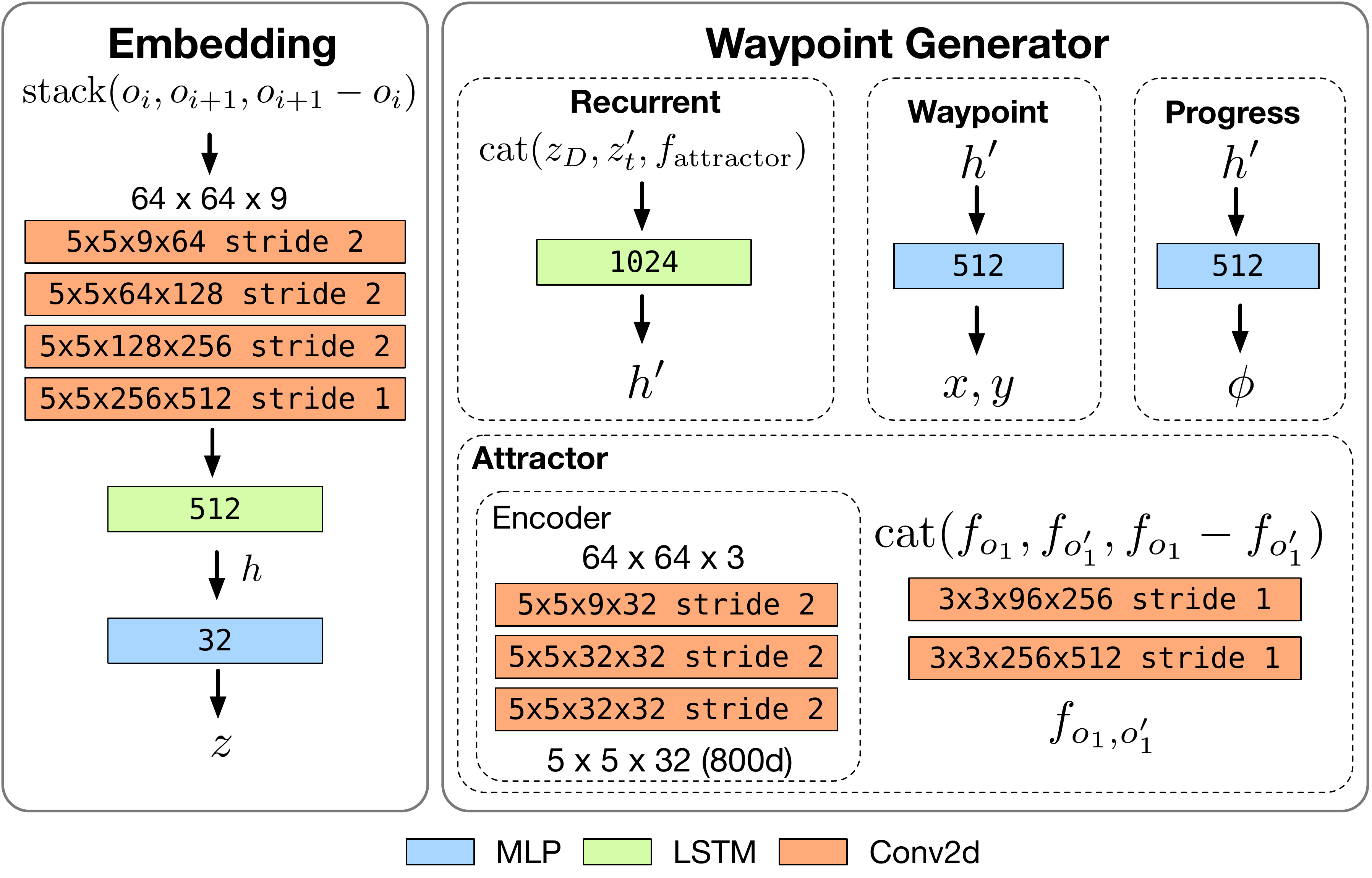}
    \caption{Neural networks used by each component in CBE. See Figure~\ref{fig:arch} for how these components work together.}
    \label{fig:networks}
    \figurespace
\end{figure}

\textbf{Network designs.} Figure~\ref{fig:networks} details the network architectures of CBE modules. The networks are lightweight (70~MB) and can run in real time on an embedded system. We use the Adam optimizer with a learning rate of 0.0003 and a learning rate decay of 0.7. Every epoch contains 200k samples. We trained CBE for 5 epochs. All baselines were also trained using the same dataset for 5 to 7 epochs.

\begin{figure}
    \centering
    \includegraphics[width=0.8\columnwidth]{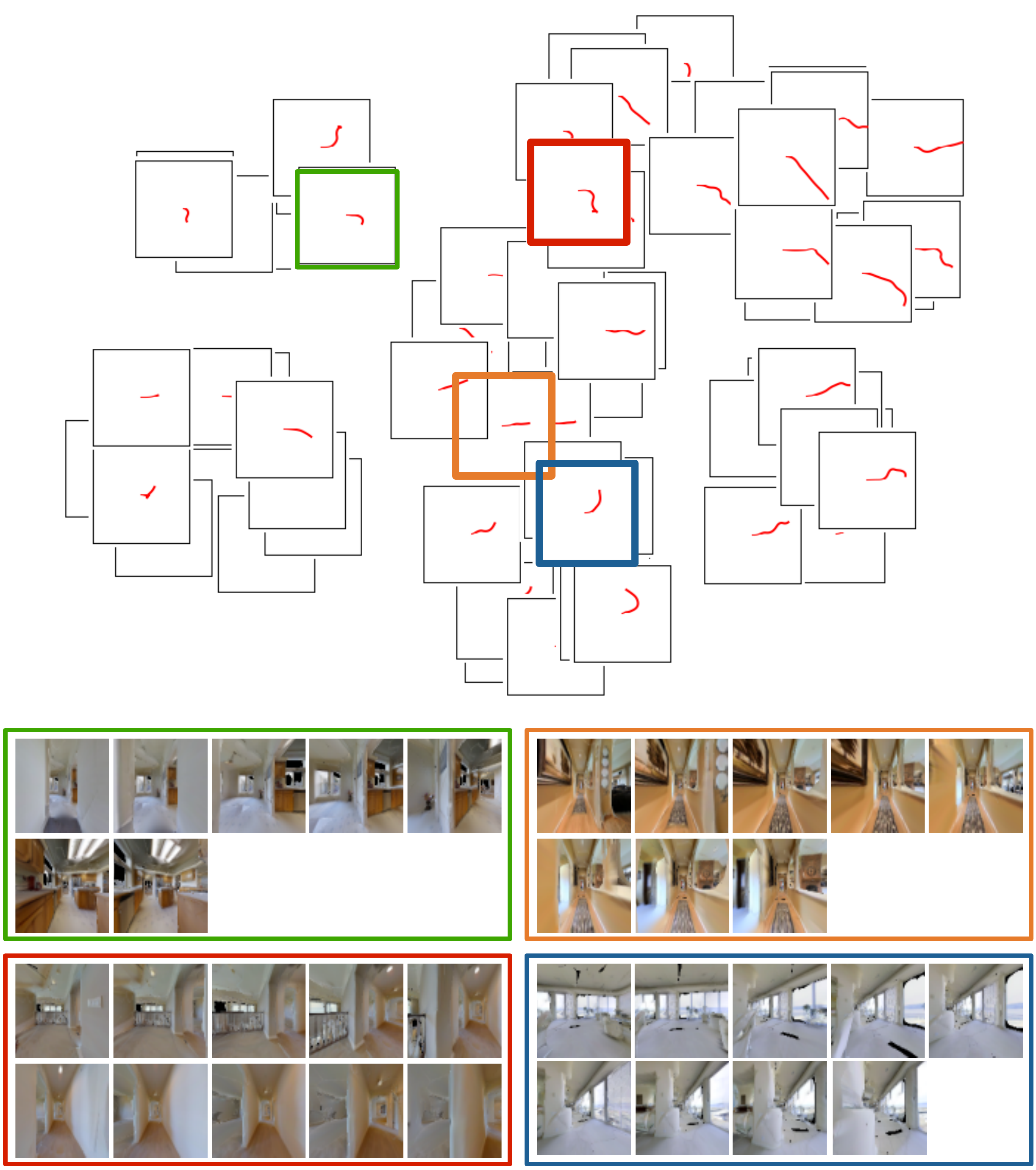}
    \caption{t-SNE visualization of the behavior manifold. Each red line visualizes an encoded trajectory. The initial pose of the robot is always at the center, pointing rightwards. 4 example trajectories are shown at the bottom.}
    \label{fig:tsne}
    \figurespace
\end{figure}

\begin{figure*}[t!]
\captionsetup[subfigure]{aboveskip=0pt}
\begin{subfigure}[b]{0.33\textwidth}
    \centering
    \includegraphics[width=1.0\textwidth]{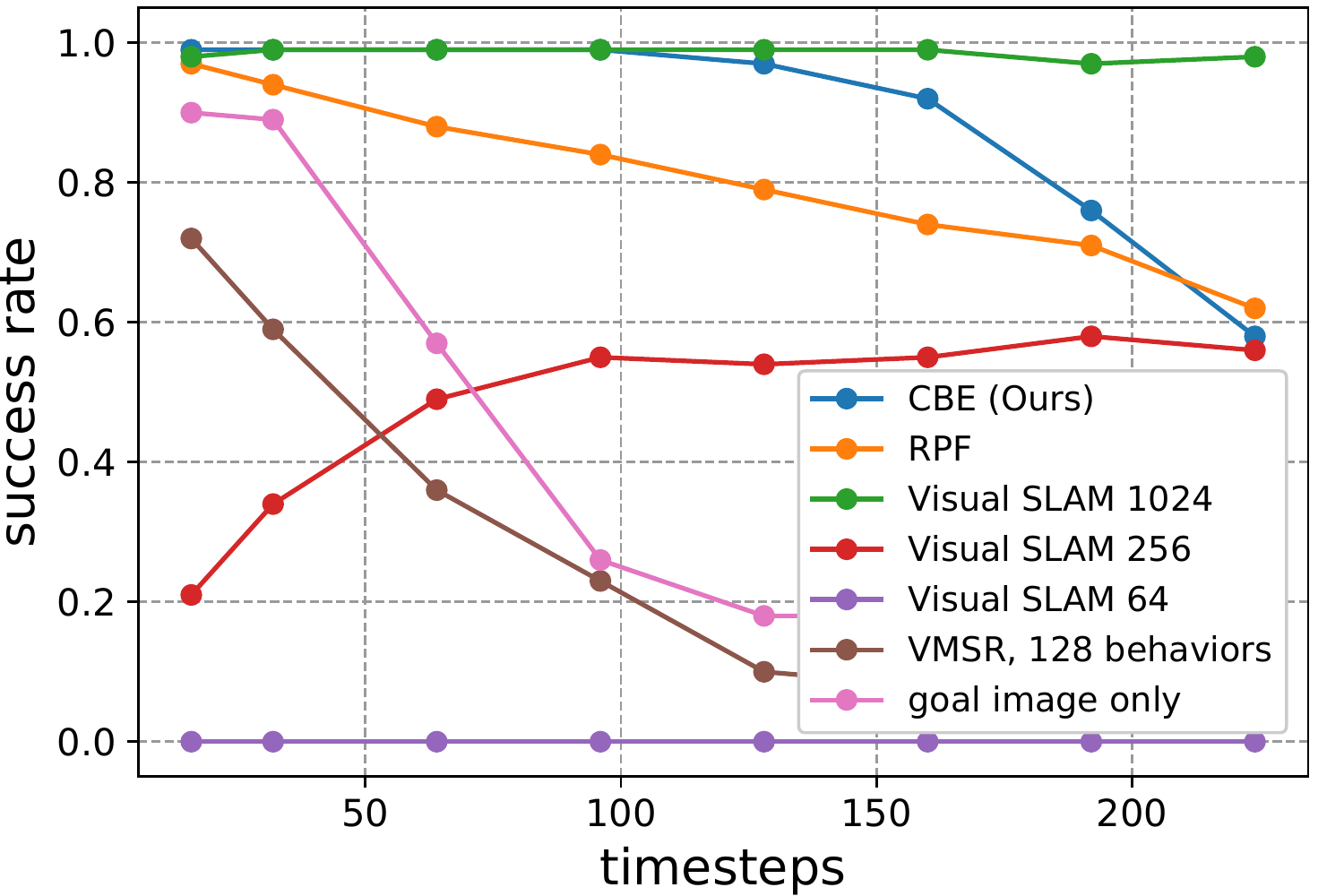}
    \caption{}
    \label{fig:single_beahvior_baselines}
\end{subfigure}\hfill
\begin{subfigure}[b]{0.33\textwidth}
    \centering
    \includegraphics[width=\textwidth]{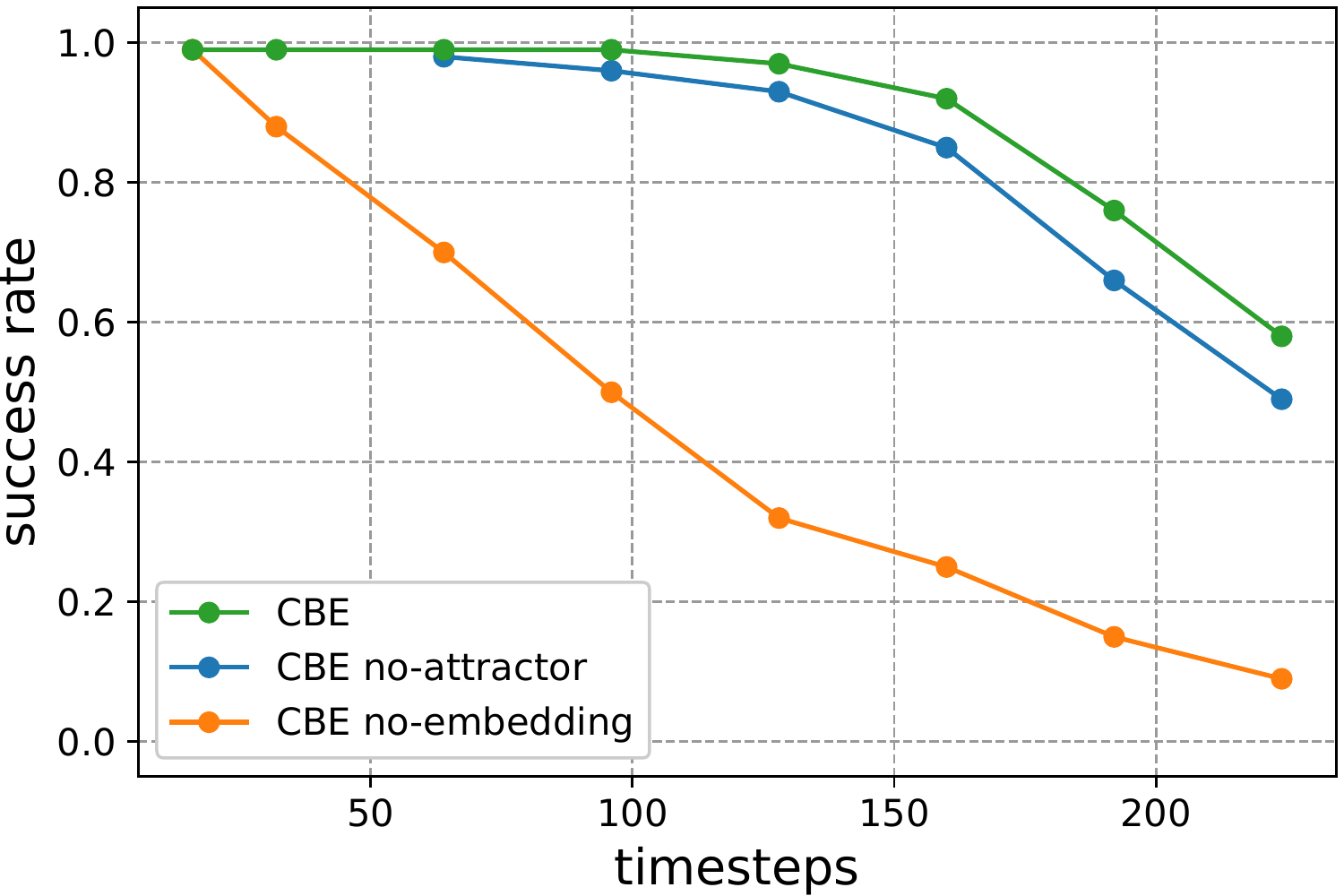}
    \caption{}
    \label{fig:visual_attractor}
\end{subfigure}\hfill
\begin{subfigure}[b]{0.33\textwidth}
    \centering
    \includegraphics[width=\textwidth]{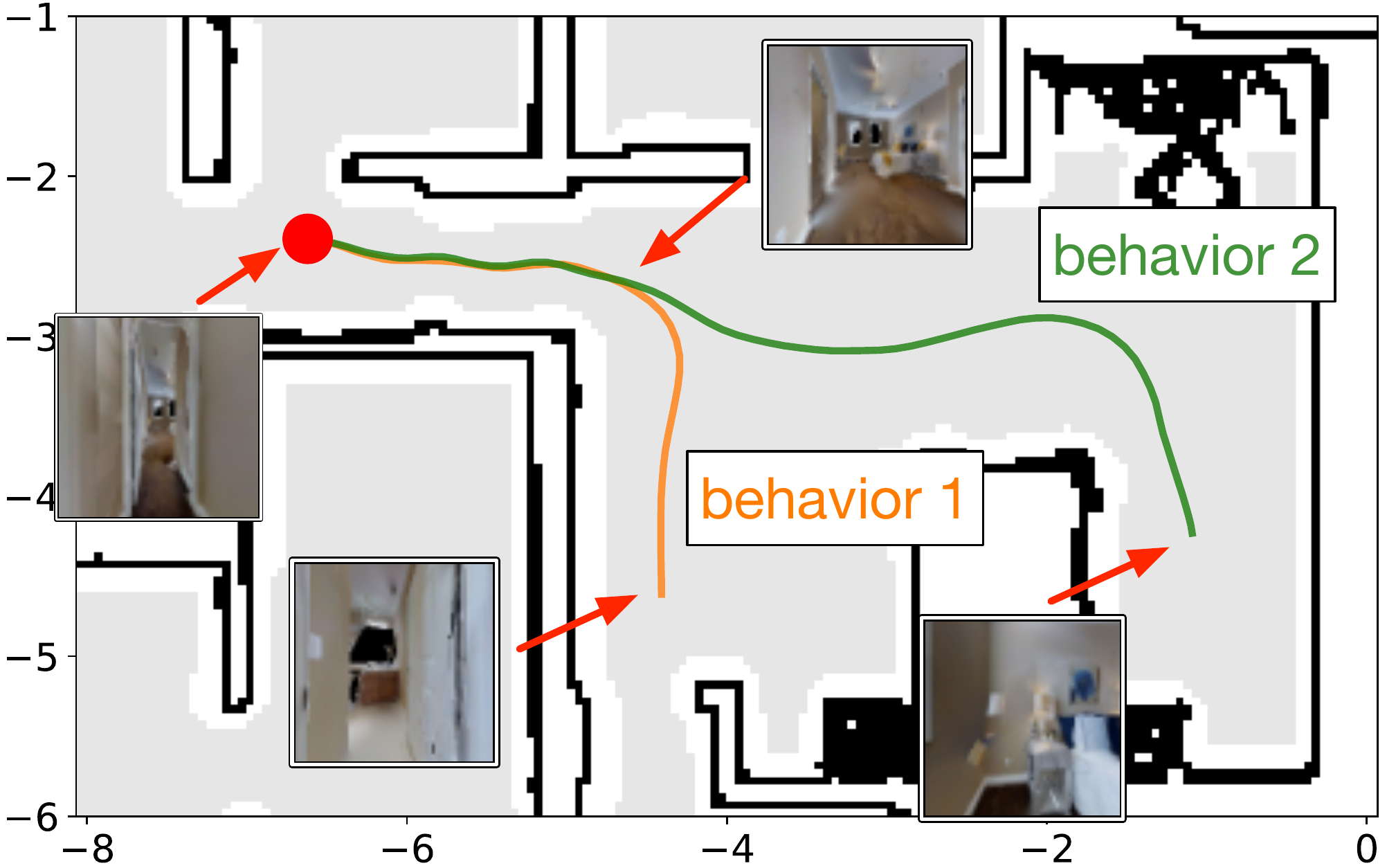}
    \vspace{-5pt}
    \caption{}
    \label{fig:embedding_execution_example}
\end{subfigure}
\caption{Evaluating CBE for single-behavior navigation. (a) Comparing CBE with baselines. The abnormal degradation of Visual SLAM 256 for short trajectories is due to initialization failures. (b) Model ablation. (c) Example rollouts of two behaviors with similar structures.}
\figurespace
\end{figure*}

\section{Experimental Results}
\label{sec:result}

\subsection{Behavior Embedding}
\label{sec:eval_embedding}

Figure~\ref{fig:tsne} shows the t-SNE plot of embeddings extracted from training trajectories. The plot shows that the embedding space encodes a meaningful behavior manifold. From left to right, trajectory lengths are increasing. From top to bottom, there is a smooth progression from ``right turns'' to ``going straight'' and to ``left turns''. The embedding space learns to encode visual odometry, even though it is not explicitly told to do so. We think this is why the learned embeddings generalize well to novel environments and can encode long-range behaviors, while being low-dimensional.

\subsection{Single-behavior Navigation}

\label{sec:eval_behavior_execution}
We study how well a robot can navigate between two locations with a single CBE behavior in unseen environments. We collected a set of trajectories of lengths ranging from 16 time steps to more than 200 time steps, with 500 trajectories collected for each time step. We extract an embedding from each trajectory to condition the waypoint generator.  We jitter the robot's initial pose to simulate imperfect alignment. We compare with the following baselines:

\paragraph{Visual SLAM} We adopt ORB-SLAM2 \cite{orbslam} which is one of the state-of-the-art real-time SLAM methods. We first feed the image sequence to reconstruct the environment and the trajectory. During execution, we run SLAM in tracking mode which localizes and tracks the pose of the robot. We set the next waypoint to be the point on the trajectory that is 5 keyframes away from the robot's current location. If localization fails, the robot will use the previously computed waypoint until localization succeeds.

\paragraph{RPF} RPF \cite{kumar2018visual} extracts a feature vector from each observation and uses attention to track the progress of a robot. Original RPF assumes the availability of camera pose and action at each time step. Here we only assume RPF has access to visual observations, same as ours.

\paragraph{VMSR} VMSR \cite{kumar2020learning} clusters fixed-length demonstrations into a discrete set of behaviors. To support variable-length trajectories, we use a recurrent encoder similar to ours instead of a convolutional encoder. Again, VMSR uses raw observations as input.

\paragraph{Goal image only} we use the local controller in \cite{meng2019scaling} because it shows strong performance when the goal image is visually reachable. We will compare \cite{meng2019scaling} against CBE in Sec.~\ref{sec:traj_following} for its path following performance.

\begin{figure}
    \centering
    \includegraphics[width=1\columnwidth]{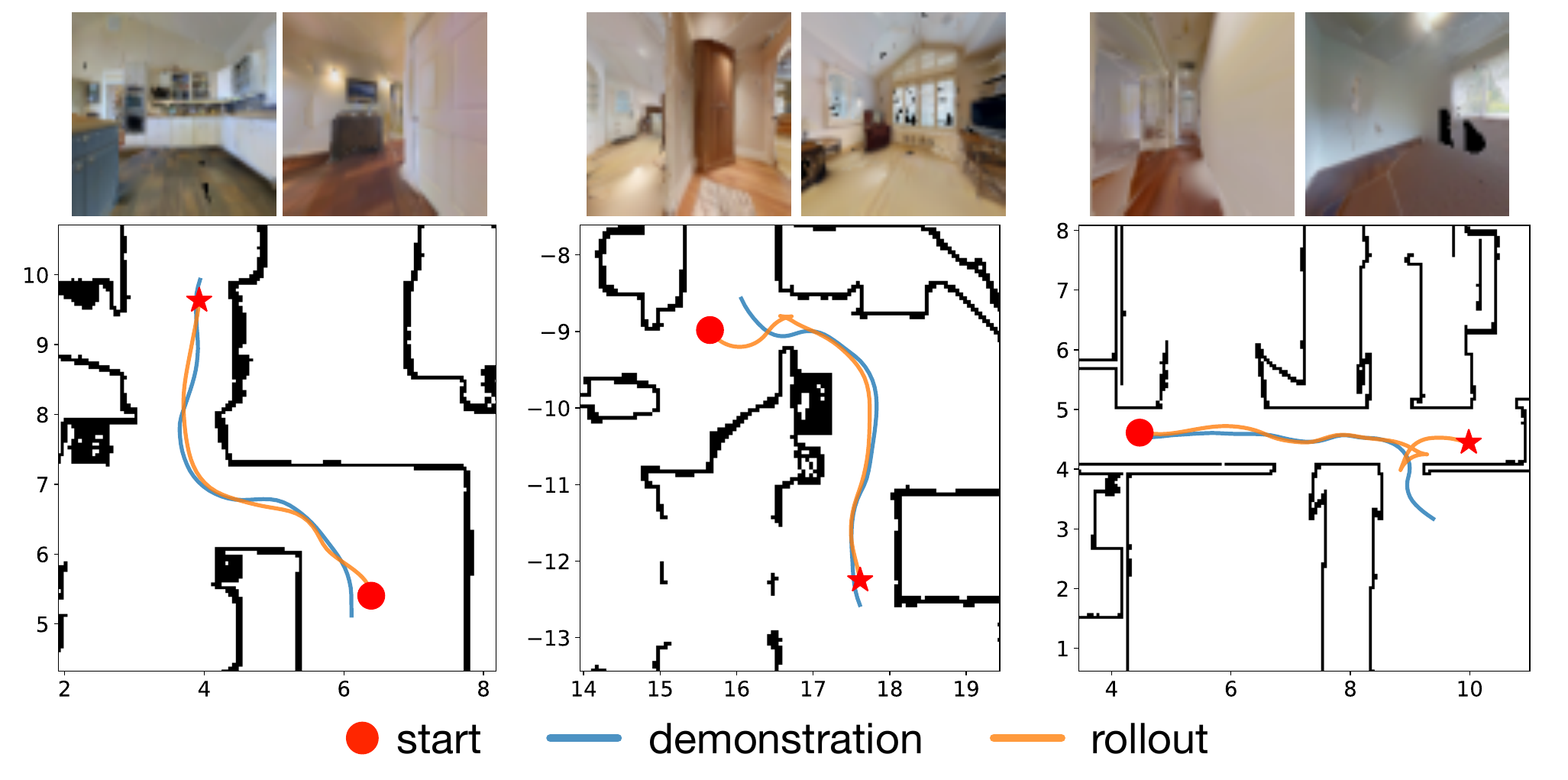}
    \caption{Example traces (including one failure case) in test environments using a single behavior embedding. Start and goal images are shown at the top. Note that starting locations of the robot are not always aligned with the beginnings of the demonstrations.}
    \label{fig:traces}
    \figurespace
\end{figure}

\begin{figure*}
    \centering
    \includegraphics[width=0.95\textwidth]{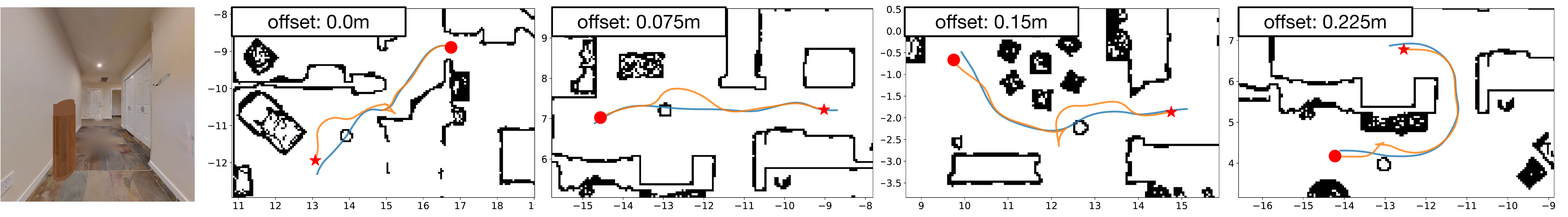}
    \caption{Handling unseen obstacles during behavior execution. Left image: robot's view of the obstacle. 4 example executions with different obstacle offsets are shown on the right. Blue trajectory: demonstration. Orange trajectory: rollout. Red dot: starting location.}
    \label{fig:obstacles}
    \vspace{-2mm}
\end{figure*}

\begin{figure*}
\captionsetup[subfigure]{aboveskip=0pt}
\begin{minipage}[b]{0.7\textwidth}
\begin{subfigure}[b]{0.49\textwidth}
    \centering
    \includegraphics[width=\textwidth]{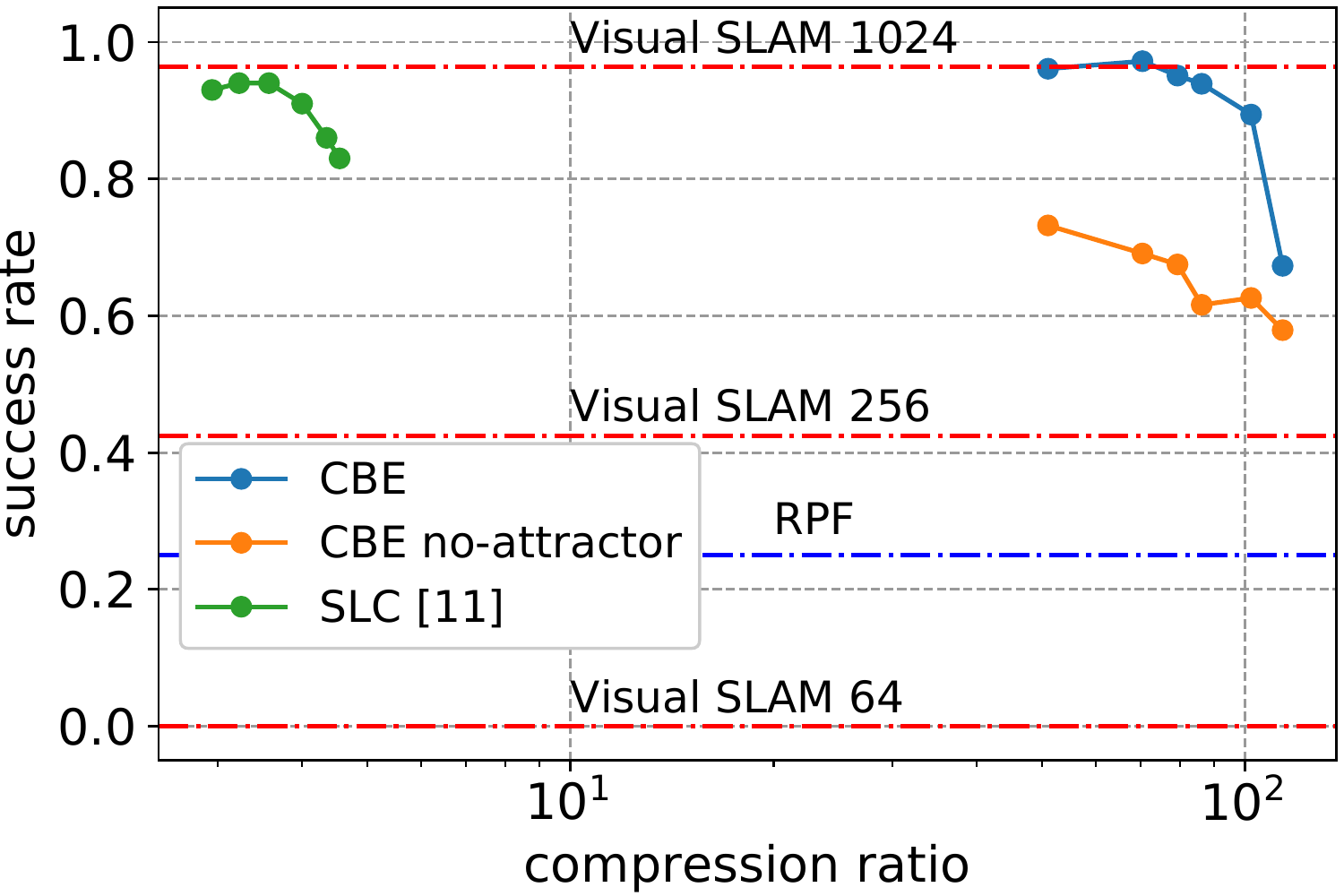}
    \caption{}
    \label{fig:path_following}
\end{subfigure}
\begin{subfigure}[b]{0.49\textwidth}
    \centering
    \includegraphics[width=0.95\textwidth]{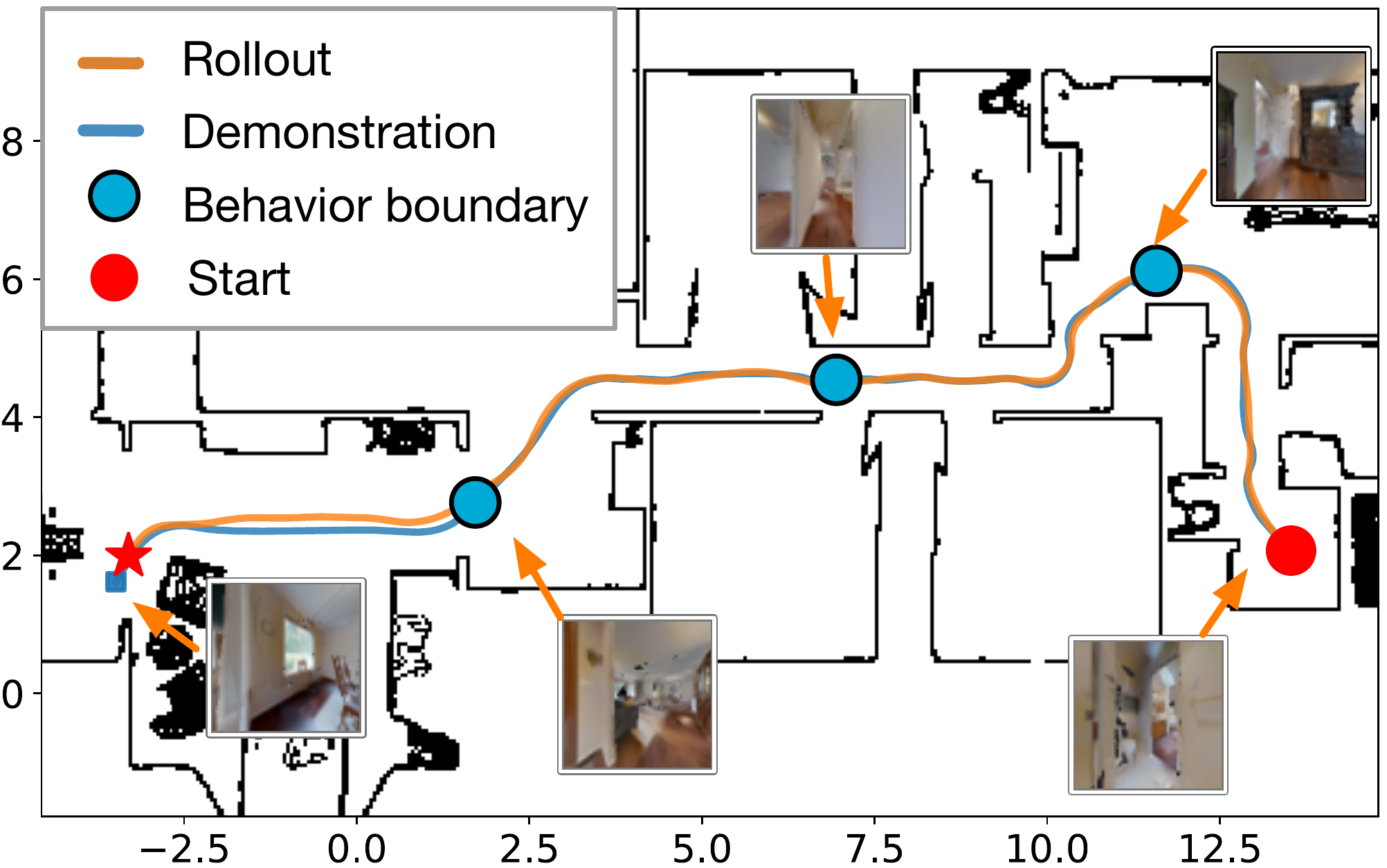}
    \vspace{4mm}
    \caption{}
    \label{fig:path_following_example}
\end{subfigure}
\caption{(a) Comparing success rates at different compression ratio. For Visual SLAM and RPF~\cite{kumar2018visual} we only show single success rates because they do not subsample observations. (b) CBE robustly follows a long path by segmenting the path into a sequence of behaviors.  See the supplementary video for more examples.}
\end{minipage}\hfill
\begin{minipage}[b]{0.28\textwidth}
    \centering
    \includegraphics[width=0.85\textwidth]{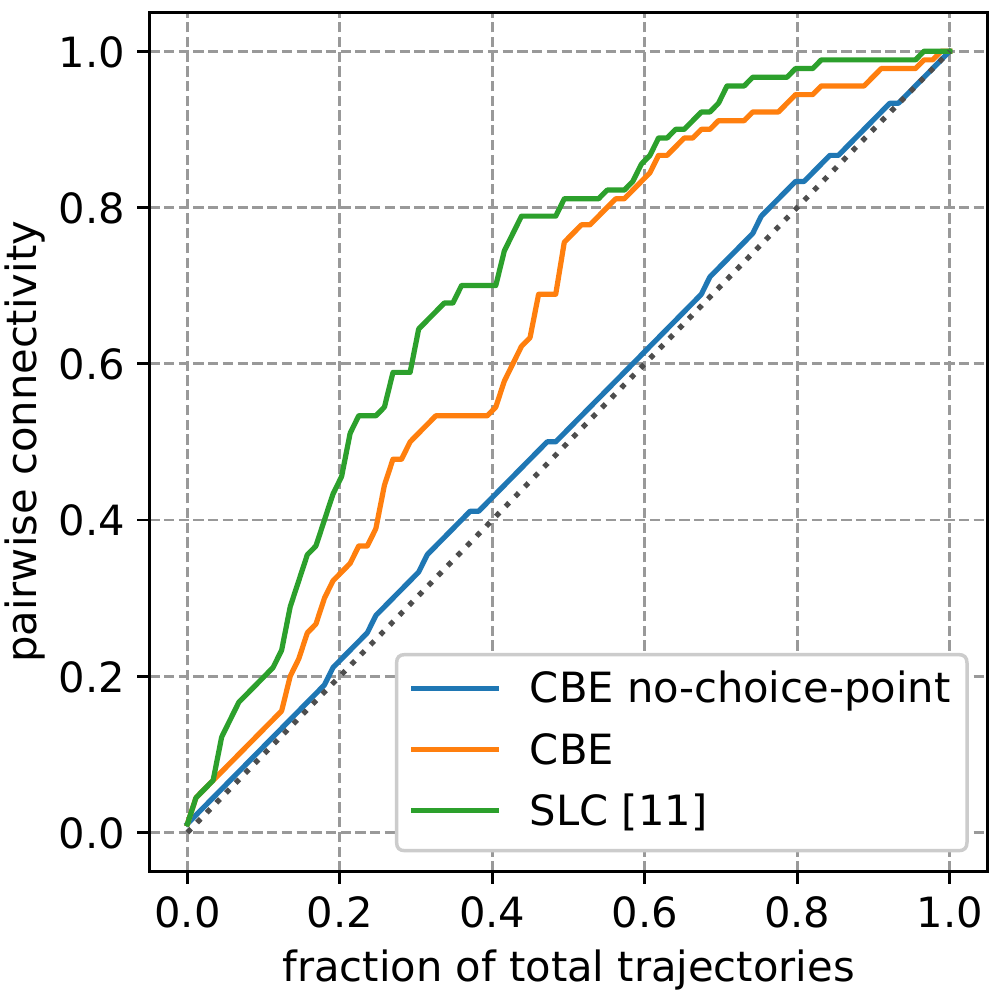}
    \caption{Comparing map connectivity when only a fraction of total trajectories are used to build the maps. Diagonal line indicates no generalization.}
    \label{fig:choice_points}
\end{minipage}
\figurespace
\end{figure*}

Figure~\ref{fig:single_beahvior_baselines} compares the success rates of CBE against the baselines (we also experimented with the SPL \cite{anderson2018evaluation} metric with almost identical results). Using goal image alone shows poor performance due to visual occlusion. CBE achieves $>95\%$ success rate for trajectories of up to 128 time steps (approx. 6~m in metric length). While CBE degrades for longer trajectories, we perform segmentation to maintain strong performance (Sec.~\ref{sec:traj_following}). RPF relies on accurate attention to track a path, but drift in attention may cause RPF to lose track and deviate from the path. VMSR clusters input trajectories into a discrete set of behaviors, hence it cannot capture the variations of behaviors well. While Visual SLAM outperforms single-embedding CBE for long trajectories, it degrades quickly as resolution decreases (degraded by 50\% with $256\times 256$ images and failed completely with $64\times 64$ images). In contrast, CBE works well with low-resolution images, and can potentially be deployed on miniature robots with fast-moving cameras.

Figure~\ref{fig:embedding_execution_example} shows how learned embeddings can distinguish between two similar behaviors. These two behaviors share the same structure: go straight and turn right. However, behavior 2 needs to go straight for a longer distance before turning right. CBE captures the difference in distance so that a robot can reach both locations with no ambiguity. Figure~\ref{fig:traces} shows more example traces.

\textbf{Model ablation}. Figure~\ref{fig:visual_attractor} compares CBE with two variants. Removing the attractor model is detrimental because it would not be able to capture the initial misalignment. This effect is more pronounced when following a sequence of behaviors (Sec.~\ref{sec:traj_following}). Removing the embedding significantly degrades performance, as the robot has to rely on the  goal attractor which can be occluded in long trajectories.

\textbf{Robustness to unseen obstacles.} To understand the robustness of our model in a dynamic environment, we randomly place a trashcan of size $0.3\times 0.3 \times 1.0~\text{m}$ close to a trajectory and let the robot execute the corresponding behavior (128 time steps). Note that the behavior is encoded when the obstacle is not present. We evaluated our model on more than 300 trajectories and the following table shows the results by varying the offset of the obstacle to trajectories:
\begin{table}[h]
  \centering
  \small
  \vspace{-1ex}
  \begin{tabular}{l|*{5}{c}}
    offset (m) &  0.0 & 0.075 & 0.15 & 0.225\\
    \hline
    Success\% & 78.8 & 82.6 & 85.1 & 89.0\\
  \end{tabular}
  \label{tab:controller_success_rate}
  \vspace{-1ex}
\end{table}

Figure.~\ref{fig:obstacles} shows example executions. The robot can successfully avoid most of the obstacles and reach the goal. The low-level controller \emph{deliberately} makes the robot deviate from the demonstration to avoid the obstacle, but since CBE uses visual feedback to follow the encoded trajectory, it can generate corrective waypoints to get the robot back on track. Note that our model is trained without obstacles. Training the model with obstacles could further improve its robustness and we leave it as future work.

\textbf{Robustness to actuation noise.} We apply a random scale $u$ to the controls of the robot at every time step. We compute $u = \text{clip}(x, -s, s) + 1.0$ where $x\sim\mathcal{N}(0.0, s/2)$. Intuitively, $s=0.5$ means that we apply a +/- 50\% random scale to the velocity and steering angle (independently). We observed 2\% and 8\% degradation at $s=0.5$ and $1.0$, respectively. This shows that our model is robust to actuation noise.

\begin{table*}
  \centering
  \small
  \begin{tabular}{l|*{2}{c}l}
    \toprule
    Method & Avg. Mem (KB) & SR & Breakdown of memory usage per trajectory (average)\\
    \hline
    CBE (Ours) & \textbf{19} & \textbf{97.2} & 6 attractors (3.2 KB each) + embeddings (32 floats = 128 bytes each)\\
    SLAM \cite{orbslam} \footnotesize{$1024\times 1024$} & 341 & 96.3 & 10933 descriptors (32 bytes each)\\
    \multicolumn{1}{r|}{\footnotesize{$256\times 256$}} & 199 & 42.5 & 6382 descriptors.\\
    \multicolumn{1}{r|}{\footnotesize{$64\times 64$}} & - & 0.0 & Failed to initialize.\\
    RPF \cite{kumar2018visual} & 424 & 21.7 & 212 feature vectors (512 floats = 2 KB each)\\
    SLC \cite{meng2019scaling} & 1548 & 93.7 & 129 images (12 KB each)\\
    \bottomrule
  \end{tabular}
  \caption{\small{Comparing memory efficiency and success rate (SR) of different methods for long-horizon visual path following. All methods use $64\times 64$ images except for SLAM which we evaluate on multiple resolutions. Note that SLAM requires extra memory to store the pose graph, which we did not include here due to the difficulty of estimating the value accurately.}}
  \label{tab:memory}
  \vspace{-3mm}
\end{table*}

\begin{table*}[t]
  \centering
  \small
  \begin{adjustbox}{width=1\textwidth}
  \begin{tabular}{l|ccc|ccc|ccc|ccc|ccc}
    \toprule
    \textbf{Envs}    
    & \multicolumn{3}{c}{\textbf{Calavo}} 
    & \multicolumn{3}{c}{\textbf{Frierson}} 
    & \multicolumn{3}{c}{\textbf{Kendall}}
    & \multicolumn{3}{c}{\textbf{Ooltewah}}
    & \multicolumn{3}{c}{\textbf{Sultan}}\\
    \#images 
    & \multicolumn{3}{c}{29,449}
    & \multicolumn{3}{c}{48,835}
    & \multicolumn{3}{c}{51,059}
    & \multicolumn{3}{c}{80,394}
    & \multicolumn{3}{c}{31,685}\\
    & \#verts & storage & SR 
    & \#verts & storage & SR
    & \#verts & storage & SR
    & \#verts & storage & SR
    & \#verts & storage & SR  \\
    \hline
    SPTM~\cite{savinov2018semi} & 3067 & 6.0 & 3.3 & 5097 & 10.0 & 0.0 & 5320 & 10.4 & 0.0 & 8287 & 16.2 & 2.2 & 3290 & 6.4 & 0.0\\
    SLC~\cite{meng2019scaling} & 617 & 36.1 & \textbf{97.8} & 935 & 54.8 & 90.4 & 805 & 47.2 & 98.1 & 1115 & 65.3 & \textbf{96.7} & 759 & 44.5 & 86.7\\
    CBE (Ours) & \textbf{357} & \textbf{1.2} & \textbf{97.8} & \textbf{498} & \textbf{1.6} & \textbf{100} & \textbf{490} & \textbf{1.6} & \textbf{100} & \textbf{611} & \textbf{2.0} & 92.3 & \textbf{388} & \textbf{1.3} & \textbf{98.9}\\
    \bottomrule
  \end{tabular}
  \end{adjustbox}
  \caption{\small{Comparing sizes of topological maps and planning success rates. Storage is in MegaBytes. SR indicates planning and trajectory following success rate. For SPTM we do a 10x subsampling of input observations. SLC does adaptive subsampling with a sparsification threshold of 0.98. For CBE we use $K=100$ for creating behavior segments.}}
  \label{tab:map_size}
  \figurespace
\end{table*}

\subsection{Long-horizon Visual Path Following} 
\label{sec:traj_following}

A common navigation task that robots perform is to navigate between two places \cite{hirose2019deep, kumar2018visual, meng2019scaling}. We show that by incorporating behaviors, a robot can follow a long trajectory with very sparse guidance. We sparsify a trajectory by segmenting it into a sequence of behaviors (Sec.~\ref{sec:segmentation}). We compare with $\cite{meng2019scaling}$ that sparsifies a trajectory by reasoning about target reachability. Compression ratio is defined as $T / N$, where $N$ is the number of landmarks and $T$ is the total number of observations. For CBE, $N$ is approximately equal to the number of behaviors. We vary  segment length $K$ in Sec.~\ref{sec:segmentation} to adjust the number of behaviors. For Visual SLAM and RPF, we only report their success rates.

We selected semantically meaningful locations (e.g., rooms) in each test environment as starts and goals and generated 500 long trajectories with an average length of 20~m. A robot follows each trajectory with a jittered initial pose. Figure~\ref{fig:path_following} compares the trajectory following success rates at different sparsity levels. Incorporating behaviors significantly increases sparsity compared to a behavior-less approach \cite{meng2019scaling}. Without visual attractors, CBE performs considerably worse, as the robot is not able to calibrate its state well to switch to the next behavior reliably. Visual SLAM performs competitively with high-resolution images, but fails completely when using the same low-resolution images as other baselines. Figure~\ref{fig:path_following_example} shows a qualitative example, where a 20~m long trajectory (450 time steps) is segmented into four behaviors and the robot executes the behaviors sequentially to reach the goal.

\textbf{Memory efficiency.} Table~\ref{tab:memory} shows that CBE is at least 10\emph{x} more efficient at encoding visual demonstrations than the baselines. A trajectory sparsified by CBE usually contains fewer than 10 embeddings (32 floats each), interleaved by visual attractors (800 floats each). In comparison, Visual SLAM stores over ten thousand feature descriptors, and existing learning-based methods require storing either dense visual features or raw images. This opens up opportunities to build compact topological maps of novel environments, studied next.

\subsection{Behavior-based Topological Mapping}
\label{sec:mapping}
We follow the same setup as in \cite{meng2019scaling, savinov2018semi}, where a robot builds a topological map of an environment from a set of experience trajectories consisting of RGB observations. A topological map is a directed graph where vertices are anchor observations selected from the trajectories and edges encode connectivity. This graph structure is often used in goal-conditioned navigation tasks, where a robot needs to plan a least-cost path to get to a specified goal.

We leverage behaviors to build sparse and well-connected topological maps. We first perform choice-point based segmentation as described in Sec.~\ref{sec:recomposition}, followed by distance-based segmentation (Sec.~\ref{sec:segmentation}) if needed. Each vertex stores an 800-dim  attractor feature. Edges are either behavioral edges (via segmentation) or proximal edges (created by linking attractors). A robot first localizes itself and the goal using a network that predicts visual overlap \cite{meng2019scaling}. Then the robot uses the Dijkstra algorithm to find the shortest path and executes the sequence of behaviors along the path.

\begin{figure}
    \centering
    \includegraphics[width=0.9\columnwidth]{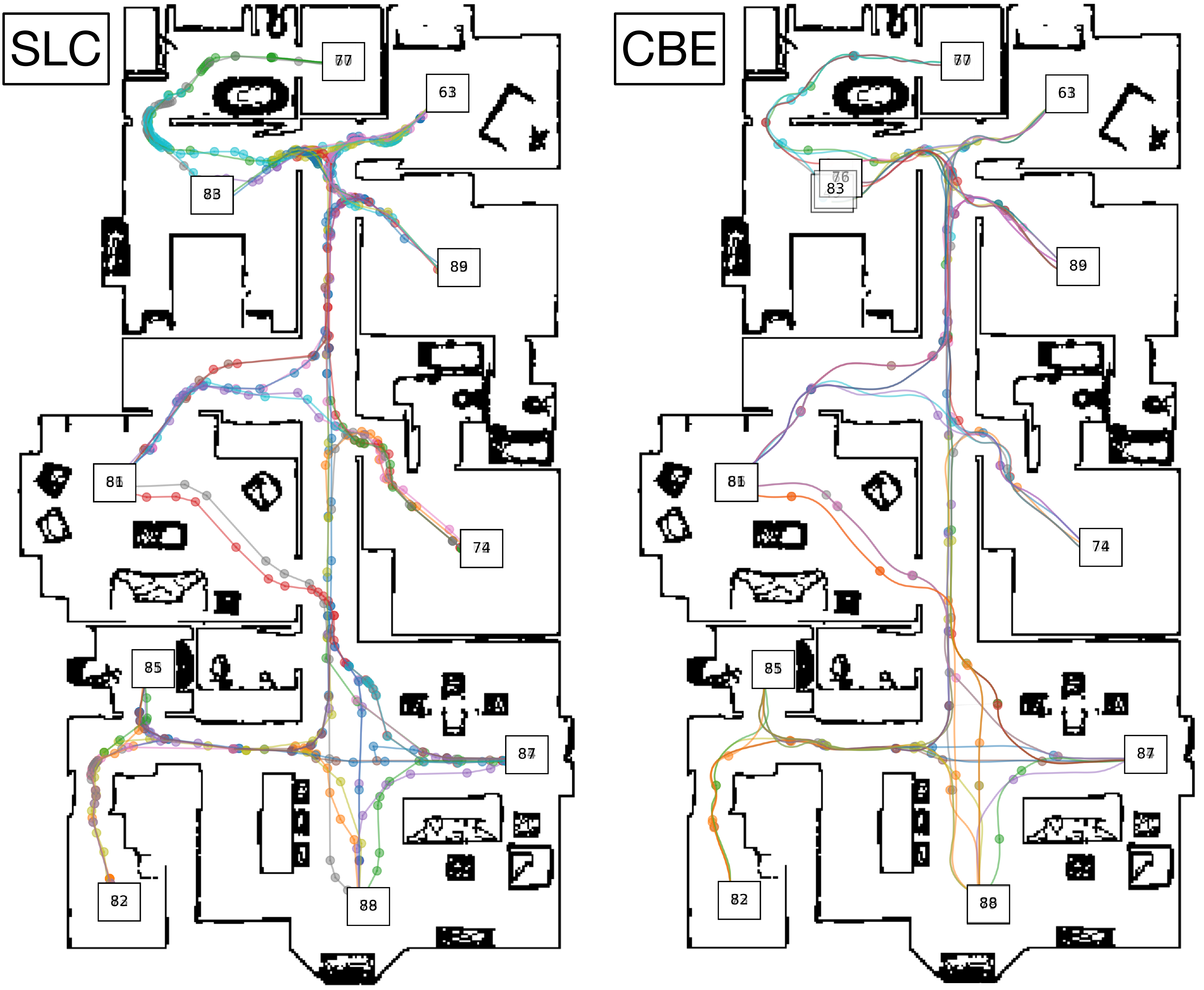}
    \caption{Visualization of the topological maps built by SLC~\cite{meng2019scaling} (no behavior) and CBE (with behaviors) in one of the test environments (Calavo). Each circle is a vertex. Each demonstration is assigned a different color. See the supplementary video for more examples.}
    \label{fig:nav_graph}
    \vspace{-6mm}
\end{figure}

We selected 10 to 14 semantically meaningful locations (e.g., rooms) in each test environment and collected pairwise trajectories to cover most of the traversable area. We built a topological map out of these trajectories. For planning, we let a robot start at one of the locations, plan a path to every other location, and follow the path. Table~\ref{tab:map_size} compares the sizes of topological maps built by CBE and planning success rates against the baseline methods. CBE builds much more compact maps and is significantly more memory efficient. SPTM~\cite{savinov2018semi} subsamples observations and extracts a 512-dim feature vector from each observation. However, it is not robust enough for controlling a non-holonomic robot in a continuous state and action space. SLC~\cite{meng2019scaling} performs competitively, but at the expense of storing significantly more information per vertex (five $64\times 64$ RGB images). The main failure cases of SLC are caused by faulty edges in the map due to visual aliasing. In comparison, CBE maps have much fewer vertices, store only an 800-dim feature per-vertex, and achieve similar or higher planning success rates due to less visual aliasing. Figure~\ref{fig:nav_graph} visualizes the distribution of vertices in the map.

\textbf{Impact of choice points on generalization.} To see the necessity of using choice points for mapping, we evaluate pairwise connectivity between locations when using a fraction of all pairwise demonstrations to build the map. In Figure~\ref{fig:choice_points}, we can see that without choice points there is almost no generalization. This is because fixed-distance segmentation (Sec.~\ref{sec:segmentation}) creates attractors that are inconsistently distributed in an environment, making it difficult to link attractors from different demonstrations. Detecting choice points significantly improves connectivity, but there is still a gap compared to a dense map. It can be a future work to improve choice point detection to close this gap.


\section{Conclusion}
\label{sec:conclusion}
We introduce Composable Behavior Embedding, a robot-agnostic behavior representation for visual navigation. With CBE, robots are able to robustly replicate visual navigation tasks using extremely compact representations; two attractor features and a low-dimensional vector per behavior.  We show how CBE can be incorporated into larger scale navigation systems for path following and topological mapping.  Here, CBE significantly improves memory-efficiency.   Our model operates in continuous state and action spaces, and we conducted experiments in realistic simulation environments.  We will test our system on a real robot once the hardware becomes accessible, but based on other work using these environments we are confident that our results will transfer well to real environments and robots.  The continuous trajectory embeddings learned by CBE are well suited to connect to similarly structured language embeddings and using our model to perform language-based visual navigation is an interesting direction for future research.

\section*{Acknowledgment}
We thank NVIDIA for generously providing a DGX used for this research via the NVIDIA Robotics Lab and the UW NVIDIA AI Lab (NVAIL).


%

\ifCLASSOPTIONcaptionsoff
  \newpage
\fi



%

\bibliographystyle{IEEEtran}
\bibliography{IEEEabrv,references.bib}

\end{document}